\documentclass{article}

 \usepackage[main, final]{neurips_2026}
\makeatletter
\renewcommand{\@notice}{}
\makeatother

\usepackage[utf8]{inputenc} 
\usepackage[T1]{fontenc}    
\usepackage{hyperref}       
\usepackage{url}            
\usepackage{booktabs}       
\usepackage{amsfonts}       
\usepackage{nicefrac}       
\usepackage{microtype}      
\usepackage{xcolor}         

\usepackage{amsmath}
\usepackage{multirow}
\usepackage{float}
\usepackage{tabularray}
\usepackage{multirow}
\usepackage{booktabs, caption, array}
\usepackage{amssymb}
\usepackage{lipsum}         
\usepackage{CJKutf8}        
\usepackage{nomencl}
\usepackage{glossaries}
\usepackage{xspace}
\usepackage{multirow}
\usepackage{graphicx}
\usepackage{bbding}
\usepackage{pifont}
\usepackage{algorithm}
\usepackage{subcaption}
\usepackage{enumitem}
\usepackage{multicol}
\usepackage{caption}
\usepackage{array}
\usepackage{fp}
\usepackage{multirow}   
\usepackage{makecell}
\usepackage{flushend}
\usepackage{cases}
\usepackage{color}
\usepackage{algpseudocode}
\usepackage{listings}
\usepackage{mathrsfs}
\usepackage{longtable}
\usepackage{colortbl}
\usepackage{bm}
\usepackage{caption}

\usepackage{stfloats}

\usepackage{varwidth} 
\setlist[itemize]{noitemsep, topsep=0pt}
\usepackage{arydshln}
\usepackage{lipsum}
\definecolor{codegreen}{rgb}{0,0.3,0.6}
\definecolor{codegray}{rgb}{0.5,0.5,0.5}

\usepackage{dashrule}

\usepackage[dvipsnames,svgnames,x11names,table]{xcolor}
\usepackage[most]{tcolorbox}
\definecolor{codegreen}{rgb}{0,0.3,0.6}
\definecolor{codegray}{rgb}{0.5,0.5,0.5}

\newcommand{\ignore}[1]{}

\definecolor{darkorange}{RGB}{255, 140, 0}
\definecolor{darkred}{RGB}{189, 20, 20}
\definecolor{lightgreen}{RGB}{145, 204, 117}
\definecolor{lightyellow}{RGB}{250, 200, 88}
\definecolor{darkyellow}{RGB}{230, 180, 75}
\definecolor{lightred}{RGB}{238, 102, 102}
\definecolor{lightblue}{RGB}{115, 192, 222}
\definecolor{myorange}{RGB}{255,127,0}
\definecolor{lightpurple}{RGB}{180, 128, 205}

\usepackage{relsize}
\usepackage{stfloats}

\usepackage{array}
\newcolumntype{C}{>{\centering\arraybackslash}p{2.8em}}

\definecolor{myorange}{RGB}{255,127,0}
\newtcolorbox{promptbox}[3][Judge Prompt]{
colback=black!5!white,
arc=5pt, 
boxrule=0.5pt,
fonttitle=\bfseries,
title=#1, 
before upper={\small}, fontupper=\fontfamily{ptm}\selectfont,
colframe=#2,
label=#3,
}

\title{Shattering the Autoregressive Curse: Dynamic  Epistemic Entropy Orchestrated Erasable Reinforcement Learning for LLMs}

\author{%
  Ziliang Wang$^{1*\ddagger}$~,
  Kang An$^{1,2*\dagger}$~,
  Faqiang Qian$^{1*}$,
    \\
  \textbf{Jialu Cai$^{1}$},
  \textbf{Cijun Ouyang$^{1}$},
  \textbf{Yuhang Wang$^{1\ddagger}$}~, 
  \textbf{Qibing Ren}$^{2\S}$~,
  \textbf{Yichao Wu}$^{1\S}$~
  \\
  $^1$SenseTime 
  $^2$Shanghai Jiao Tong University\\
  \texttt{\{wangziliang1,wangyuhang,wuyichao\}@sensetime.com}\\
  \texttt{ankang@gml.ac.cn, renqibing@sjtu.edu.cn}
}

\makeatletter
\renewcommand{\@thanks}{%
  \footnotetext[1]{%
    Equal contribution\hspace{1em}%
    $^{\dagger}$Work done during internship at SenseTime\hspace{1em}%
    $^{\ddagger}$Project leader\hspace{1em}%
    $^{\S}$Corresponding author%
  }%
}
\makeatother

\begin{document}

\maketitle

\begin{abstract}

Although reinforcement learning (RL) has expanded the cognitive boundaries of large language models (LLMs), it often remains vulnerable to the autoregressive curse in long-horizon logical reasoning: small epistemic perturbations introduced early in generation can propagate irreversibly along the Markov decision process flow, triggering cascading failures that drive the reasoning trajectory toward collapse. To overcome this autoregressive cascade, in which a single early mistake can compromise all subsequent reasoning steps, we propose dynamic epistemic entropy orchestrated erasable reinforcement learning ($\text{E}^3\text{RL}$). $\text{E}^3\text{RL}$ eliminates reliance on external signals by grounding the model’s endogenous local autoregressive cross-entropy as an intrinsic coordinate of epistemic uncertainty. By introducing segment-level adaptive dynamic thresholds and advantage allocation, $\text{E}^3\text{RL}$ enables the model to precisely excise localized logical defects while reusing historical key-value (KV) cache streams, thereby endowing the reasoning process with a self-healing capability. We train E³RL on the DeepMath-103k dataset. Experimental results show that $\text{E}^3\text{RL}$ reshapes the exploration efficiency of long-sequence reasoning and improves sample efficiency while maintaining linear memory overhead. On mathematical reasoning benchmarks such as AIME, $\text{E}^3\text{RL}$ achieves substantial performance gains, with the 4B and 8B parameter models surpassing previous state-of-the-art (SOTA) results by 5.349\% and 6.514\%, respectively. These findings suggest that $\text{E}^3\text{RL}$ shatters the autoregressive curse in long-sequence reasoning and establishes a theoretical and systems-level foundation for the next generation of self-healing artificial general intelligence (AGI).

\begin{figure*}[h]
\centering
\includegraphics[width=0.96\linewidth]
{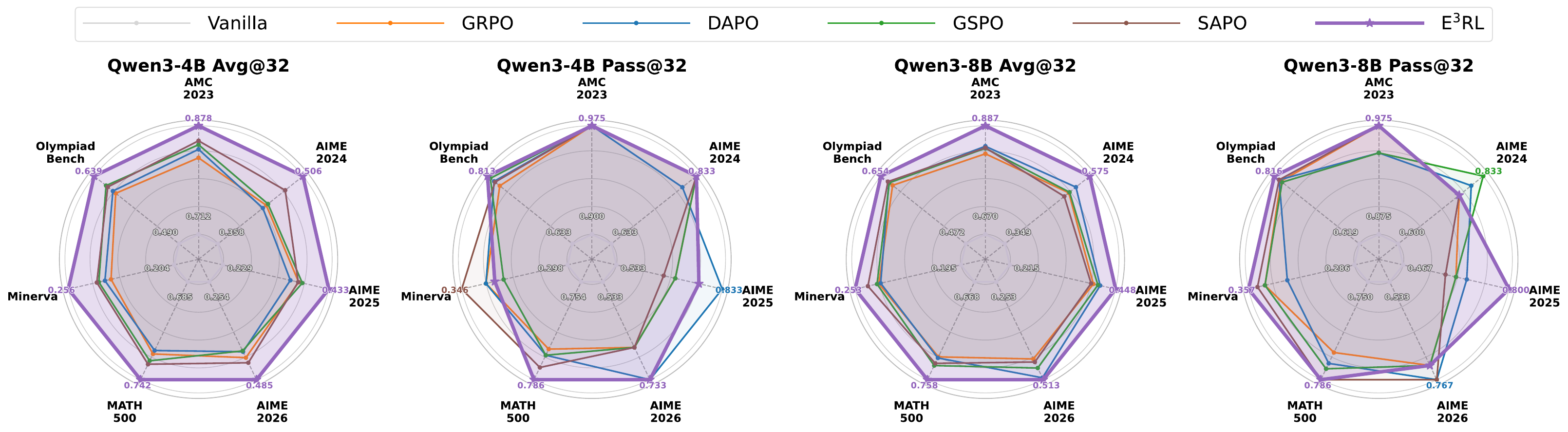}
\caption{Performance of different RL strategies.}
\label{fig:rl_strategy_radar}
\end{figure*}

\end{abstract}

\section{Introduction}


Propelled by the reinforcement learning\citep{liu2025reinforcement,zhang2025survey} driven autoregressive generation paradigm\citep{wang2026multimodal}, large language models (LLMs) map complex problem-solving into highly coherent token sequences prediction, achieving a substantial leap in cognitive reasoning capabilities\citep{zhang2026consistent,wan2026srpo,cheng2026revisiting}. However, as reasoning trajectories extend into deep and long contexts\citep{wang2025offline11, chen2025reinforcement11}, this unidirectional generation paradigm\citep{hou2025thinkprune11}, which strictly follows the temporal causal arrow\citep{chen2026learning}, exposes a critical underlying structural defect. We refer to this defect as the autoregressive curse. Devoid of spatiotemporal rollback and local error-correction operators, early-stage epistemic perturbations\citep{song2026large} are unconditionally accepted and exponentially amplified along the Markov decision flow\citep{zhu2026edis}. This amplification ultimately and inevitably drives the entire high-dimensional reasoning trajectory into a catastrophic cascading collapse\citep{shen2026reasoning, gao2026unlocking}.

Both academia\citep{zhu2024yulan,yang2026incoder} and industry\citep{team2026kimi,yang2026iquest} have long been trapped in an exceedingly costly technical misconception by attempting to utilize external systems to patch the foundational unidirectional generation defect of LLMs\citep{zheng2025stepsearch11, wang2025erase11}. Introducing an external process reward model (PRM)\citep{lightman2024let, zheng2025survey, pronesti2026beyond} for step-by-step scoring is not only limited by exorbitant data annotation costs but also, when confronting the dynamically evolving state space of large models, static evaluation networks are inevitably trapped in severe distribution shifts, consequently triggering system-level reward hacking\citep{tiwari2026reward, wang2026reward11}. Alternatively, global resampling methods\citep{kobayashi2026flexible} must wait until the complete generation of the long-tail sequence to execute discrimination and rejection\citep{dalal2026more}. This approach indiscriminately flushes entire tensor sequences containing numerous correct prefixes solely due to isolated local computational deviations\citep{yu2026curiosity}. This indiscriminate flushing not only disrupts the fine-grained credit assignment mechanisms\citep{guo2026segment, fang2026proximity, yang2026int} of reinforcement learning but also induces an exponential explosion in computational and memory overhead\citep{kim2026spend, zhang2026resource}, rendering efficient scaling on hundred-billion-parameter models infeasible.

To shatter the autoregressive curse, we propose dynamic epistemic entropy orchestrated erasable reinforcement learning ($\text{E}^3\text{RL}$). By introducing a non-Markovian erasure operator, this method restructures the gradient flow pathways of reinforcement learning. $\text{E}^3\text{RL}$ entirely eliminates external dependencies by physicalizing the endogenous local autoregressive cross-entropy\citep{cui2025entropy, wang2026beyond} of the model into coordinates that explicitly represent epistemic uncertainty. By introducing segment-level adaptive dynamic thresholds and advantage allocation, $\text{E}^3\text{RL}$ enables the model to precisely excise localized logical defects while reusing historical key-value (KV) cache streams\citep{liu2026adaptive, chen2026arborkv}. In this way, it prevents defective logic from propagating through the reasoning chain and avoids inference deadlocks. Trained on the DeepMath-103k\citep{he2025deepmath} dataset, experimental results demonstrate that $\text{E}^3\text{RL}$ effectively optimizes the exploration efficiency of long-sequence reasoning and significantly improves sample efficiency. While maintaining an exceptionally low linear memory footprint, the proposed method achieves substantial performance improvements across competitive benchmarks including AMC 2023\citep{amc2023_12a_aops, amc2023_12b_aops}, AIME 2024\citep{aime2024_i_aops, aime2024_ii_aops}, AIME 2025\citep{aime2025_i_aops, aime2025_ii_aops}, AIME 2026\citep{aime2026_i_aops, aime2026_ii_aops}, Math500\citep{lightman2023let}, Minerva\citep{lewkowycz2022solving}, and OlympiadBench\citep{he2024olympiadbench}. Notably, models at the 4B and 8B parameter\citep{yang2025qwen3} scales outperform previous state-of-the-art results by 5.349\% and 6.514\%, respectively. These findings indicate that $\text{E}^3\text{RL}$ shatters the autoregressive curse inherent in long-sequence reasoning, thereby establishing a theoretical and systemic cornerstone for next-generation artificial general intelligence (AGI).


\section{Preliminary}

Let $\mathcal{V}$ denote a finite discrete vocabulary with cardinality
$|\mathcal{V}|=V$. Given an input prompt
$x \in \mathcal{V}^{*}$, an autoregressive language model
$p_{\theta}$ maps the prompt to an output sequence
$y=(y_1,y_2,\ldots,y_T)\in\mathcal{V}^{T}$, where
$T\in\mathbb{N}^{+}$ denotes the sequence length and
$\theta\in\Theta\subseteq\mathbb{R}^{d_{\theta}}$ denotes the model parameters\citep{ji2026survey}. The conditional distribution over the output sequence follows the standard
autoregressive factorization\citep{jafari2025closed}:
\begin{equation}
p_{\theta}(y\mid x)
=
\prod_{t=1}^{T}
p_{\theta}(y_t\mid y_{<t},x),
\qquad
y_{<t}:=(y_1,\ldots,y_{t-1}).
\end{equation}

At each generation step $t$, the conditional distribution is obtained from
the hidden state $h_t\in\mathbb{R}^{d_h}$ via a softmax layer\citep{zheng2025mmrpt, zhou2026look}:
\begin{equation}
p_{\theta}(y_t=v\mid y_{<t},x)
=
\frac{
\exp\left(h_t^{\top}e_v/\gamma\right)
}{
\sum_{v'\in\mathcal{V}}
\exp\left(h_t^{\top}e_{v'}/\gamma\right)
},
\qquad
\forall v\in\mathcal{V},
\end{equation}
where $e_v\in\mathbb{R}^{d_h}$ is the output embedding of token $v$,
$\gamma>0$ is the temperature parameter, and the hidden state is computed by
a Transformer decoder\citep{su2026learning}:
\begin{equation}
h_t = \mathrm{Transformer}_{\theta}(x,y_{<t}).
\end{equation}

Equivalently, the model can be viewed as a stochastic policy
$\pi_{\theta}(\cdot\mid x,y_{<t})$ over the vocabulary\citep{qian2025uniapl}. Once a token $y_t$ is
sampled or decoded, it is irrevocably appended to the prefix and becomes part
of the conditioning context for all subsequent decisions\citep{kim2026early}. This causal
commitment is the defining property of autoregressive generation, but it also
constitutes its main structural vulnerability: an early local error cannot be
rolled back by the model itself and may therefore propagate through the
remaining reasoning trajectory\citep{huang2026not}.

To formalize this effect, let
\begin{equation}
y_t^{\star}
=
\arg\max_{v\in\mathcal{V}}
p_{\theta}(v\mid y_{<t},x)
\end{equation}
denote the greedy decoding output at step $t$, and let $y_t\neq y_t^{\star}$
be the actually sampled token\citep{jin2026corefine}. We define the cognitive perturbation
$\epsilon_t\geq 0$ as the log-probability gap between the locally optimal
token and the sampled token:
\begin{equation}
\epsilon_t
:=
\log p_{\theta}(y_t^{\star}\mid y_{<t},x)
-
\log p_{\theta}(y_t\mid y_{<t},x)
\geq 0.
\end{equation}

When $\epsilon_t=0$, the model selects a optimal token under its own
current distribution. When $\epsilon_t>0$, the model deviates from the  optimal action, and the magnitude of $\epsilon_t$ quantifies the degree of
local decision mismatch\citep{xu2026unveiling}. If such a perturbation exceeds a critical level, the
subsequent generation process is forced to continue from a cognitively biased
prefix\citep{venhoff2025understanding}.

The harmfulness of this perturbation lies in its non-compressible cascading
effect\citep{zheng2026pilot, jin2026himac}. Due to the multiplicative structure of the autoregressive chain, local
log-probability gaps accumulate additively in log space and therefore induce
exponential distortion in probability space\citep{cao2026diffcot, you2025probabilistic}. For a prefix of length $k$, the
cumulative perturbation can be written as
\begin{equation}
E_{1:k}
=
\sum_{t=1}^{k}\epsilon_t .
\end{equation}

Consequently, even a small perturbation at an early step may affect all later
conditional distributions:
\begin{equation}
p_{\theta}(y_{t+1}\mid y_{\leq t},x),
\;
p_{\theta}(y_{t+2}\mid y_{\leq t+1},x),
\ldots,
p_{\theta}(y_T\mid y_{<T},x).
\end{equation}

This observation motivates a segment-level generation and correction
mechanism. Instead of treating the entire output as a single irreversible
decision chain\citep{sharma2026prism, singh2026v_1}, we introduce intermediate checkpoints that allow the model to
monitor uncertainty, identify locally unstable reasoning segments, and erase
or resample problematic segments before they contaminate the full trajectory.

\section{Method}
\label{sec:method}
\subsection{Segmented Generation}

$E^3\mathrm{RL}$ restructures conventional one-shot autoregressive generation
into an iterative segmented generation process with checkpoints. Given an
input prompt $x$ and a pair of hyperparameters $(L,N)$, the output sequence
$y=(y_1,\ldots,y_T)$ is divided into $N$ non-overlapping segments. Assume $T=N\times L$, the $n$-th segment is defined as
\begin{equation}
s_n
=
\left(
y_{(n-1)L+1},
\ldots,
y_{nL}
\right),
\qquad
n=1,\ldots,N .
\end{equation}

Let $s_{<n}:=(s_1,\ldots,s_{n-1})$ denote all previously generated segments.
Under the segmented formulation, the probability of the $n$-th segment is
\begin{equation}
p_{\theta}(s_n\mid x,s_{<n})
=
\prod_{t=(n-1)L+1}^{nL}
p_{\theta}(y_t\mid y_{<t},x).
\end{equation}

The segmented generator decomposes the original length-$T$ irreversible
decision chain into $N$ local decision windows. This design reduces the depth
of single-step error propagation and introduces explicit checkpoints at which
the model can evaluate whether the current segment should be accepted,
erased, or regenerated.

\begin{figure*}[t]
\centering
\includegraphics[width=0.96\linewidth]
{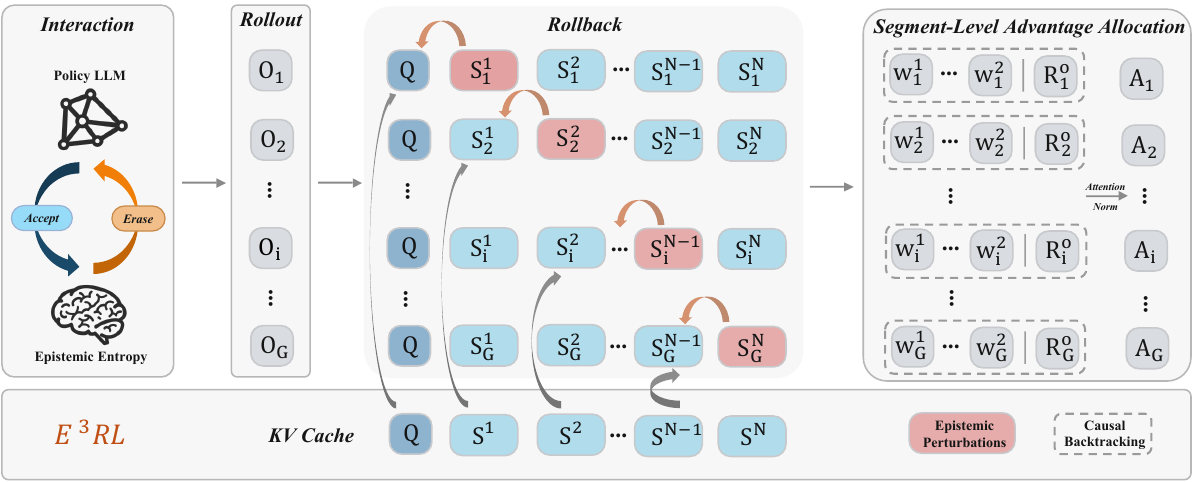}
\caption{Overview of $\text{E}^3\text{RL}$. }
\label{fig:e3rl_overview}
\end{figure*}

\subsection{Epistemic Entropy Monitoring}

For a generation position $t$, given the hidden state $h_t$, we define the
token-level cognitive entropy and its segment-level average as
\begin{equation}
H_t = -\sum_{v\in\mathcal{V}} p_{\theta}(v\mid h_t) \log p_{\theta}(v\mid h_t),\quad
\mathcal{H}_n = \frac{1}{L}\sum_{t=(n-1)L+1}^{nL} H_t .
\end{equation}

To suppress short-term entropy fluctuations and emphasize persistent
uncertainty trends, $E^3\mathrm{RL}$ introduces sliding-window smoothing
and its boundary normalization:
\begin{equation}
\widetilde{\mathcal{H}}_n = \frac{1}{C_n}\sum_{w=-W}^{W}\alpha^{|w|}\mathcal{H}_{n+w},\quad
C_n = \sum_{w=-W}^{W}\alpha^{|w|}\mathbf{1}[1\leq n+w\leq N],
\end{equation}
where $W$ is the half-window size, $\alpha\in(0,1]$ is a distance-decay
coefficient, and $C_n$ is the normalization factor near segment boundaries. In addition to the average entropy, we extract the maximum entropy within
the segment and monitor the intra-segment entropy variation rate to capture
local burst-like cognitive crises and sharp oscillations:
\begin{equation}
\mathcal{H}_n^{\max} = \max_{t\in\{(n-1)L+1,\ldots,nL\}} H_t,\quad
\Delta \mathcal{H}_n = \frac{1}{L-1}\sum_{t=(n-1)L+1}^{nL-1}\left|H_{t+1}-H_t\right|.
\end{equation}
A large value of $\Delta\mathcal{H}_n$ indicates sharp cognitive oscillation
within the segment. Such instability may reveal local reasoning defects even
when the average entropy remains moderate. Finally, the comprehensive uncertainty metric of the $n$-th segment is
defined as
\begin{equation}
U_n
=
\underbrace{\widetilde{\mathcal{H}}_n}_{\text{Base Uncertainty}}
+
\underbrace{\lambda_G \Delta\mathcal{H}_n}_{\text{Gradient Anomaly}}
+
\underbrace{
\lambda_M
\sigma_{\mathrm{sig}}
\left(
\frac{
\mathcal{H}_n^{\max}-\mu_E
}{
\sigma_E+\varepsilon_E
}
\right)
}_{\text{Extremum Deviation}},
\end{equation}
where $\lambda_G$ and $\lambda_M$ are weighting coefficients,
$\sigma_{\mathrm{sig}}(\cdot)$ denotes the sigmoid function, $\mu_E$ and
$\sigma_E$ are the mean and standard deviation of cognitive entropy in the
current batch, and $\varepsilon_E>0$ is a small constant for numerical
stability. The base term $\widetilde{\mathcal{H}}_n$ measures the overall
uncertainty of the segment, the gradient term $\Delta\mathcal{H}_n$ detects
cognitive instability, and the extremum term $\mathcal{H}_n^{\max}$ identifies
local burst-like uncertainty spikes.

\subsection{Segment-Level Advantage Assignment}

Under the group-sampling framework of GRPO, each question $q$ is associated
with $G$ sampled output sequences and their segment-level uncertainty
metrics:
\begin{equation}
\{y^1,y^2,\ldots,y^G\}\sim\pi_{\theta_{\mathrm{old}}}(\cdot\mid q),\quad
\mathcal{D}_n^{\mathrm{group}}=\left\{U_n^1,U_n^2,\ldots,U_n^G\right\}.
\end{equation}

We compute the group mean and standard deviation as
\begin{equation}
\mu_n^{\mathrm{group}}=\frac{1}{G}\sum_{i=1}^{G}U_n^i,\quad
\sigma_n^{\mathrm{group}}=\sqrt{\frac{1}{G}\sum_{i=1}^{G}\left(U_n^i-\mu_n^{\mathrm{group}}\right)^2}.
\end{equation}

The macro-level dynamic baseline and its adaptive offset are then defined as
\begin{equation}
\beta_n^{\mathrm{macro}}=\mu_n^{\mathrm{group}}+\kappa\!\left(\sigma_n^{\mathrm{group}}\right)\sigma_n^{\mathrm{group}},\quad
\kappa(\sigma)=\kappa_0+\kappa_1\tanh\left(\frac{\sigma-\sigma_0}{\sigma_0+\varepsilon_{\sigma}}\right).
\end{equation}
This offset function adaptively relaxes or tightens the rejection threshold
according to the variance of group-level entropy, which reflects the ambiguity
of the current problem. At the micro level, for the $e_n$-th erasure attempt of the $n$-th segment,
we introduce an exponentially increasing penalty factor and the final adaptive
threshold:
\begin{equation}
\Gamma(e_n)=\exp\left(\eta e_n^{\delta}\right),\quad
\Theta_n(e_n)=\beta_n^{\mathrm{macro}}\cdot\Gamma(e_n)\cdot\phi(\mathcal{H}_{<n}),
\end{equation}
where $\eta>0$ controls the penalty strength and $\delta>0$ controls the
growth order. The history-dependent modulation function is
\begin{equation}
\phi(\mathcal{H}_{<n})=
1+\rho\tanh\left(
\frac{1}{n-1}\sum_{m=1}^{n-1}
\frac{\widetilde{\mathcal{H}}_m-\beta_m^{\mathrm{macro}}}{\beta_m^{\mathrm{macro}}+\varepsilon_{\beta}}
\right),\qquad n>1.
\end{equation}
$E^3\mathrm{RL}$ extends GRPO from sequence-level optimization to
segment-level optimization. Given the final sequence-level reward $R(y^i)$,
we use causal backtracking assignment to obtain the reward signal for the
$n$-th segment. The token attribution weight, the segment-level reward, and
its normalization factor are defined as
\begin{equation}
a_t^i=\frac{1}{L'}\sum_{t'=T-L'+1}^{T}\mathrm{Attn}_{t'\rightarrow t}^{i},\quad
R_n^i=\frac{R(y^i)}{Z^i}\sum_{t=(n-1)L+1}^{nL}a_t^i,\quad
Z^i=\sum_{\tau=1}^{T}a_{\tau}^i,
\end{equation}
where $L'$ is the length of the terminal attribution window and
$\mathrm{Attn}_{t'\rightarrow t}^{i}$ denotes the attention mass from token
$t'$ to token $t$ in the $i$-th trajectory. The segment-level advantage of the $n$-th segment in the $i$-th sequence is
defined by group normalization:
\begin{equation}
A_n^i=
\frac{
R_n^i-\frac{1}{G}\sum_{j=1}^{G}R_n^j
}{
\sqrt{
\frac{1}{G}\sum_{j=1}^{G}\left(
R_n^j-\frac{1}{G}\sum_{k=1}^{G}R_n^k
\right)^2
}+\varepsilon_A
}.
\end{equation}

The segmented policy optimization objective of $E^3\mathrm{RL}$ is formulated as
\begin{equation}
\begin{aligned}
J_{E^3\mathrm{RL}}(\theta)=
\mathbb{E}_{q\sim P(\mathcal{Q}),\{y^i\}_{i=1}^{G}\sim\pi_{\theta_{\mathrm{old}}}}
\Bigg[
\frac{1}{G}\sum_{i=1}^{G}\sum_{n=1}^{N}
\min\Big(
\rho_n^i(\theta)A_n^i,\\
\mathrm{clip}\left(\rho_n^i(\theta),1-\epsilon,1+\epsilon\right)A_n^i
\Big)
-\beta D_{\mathrm{KL}}\left(\pi_{\theta}\Vert\pi_{\mathrm{ref}}\right)
\Bigg],
\end{aligned}
\end{equation}
where the segment-level probability ratio is
\begin{equation}
\rho_n^i(\theta)=
\frac{\pi_{\theta}(s_n^i\mid q,s_{<n}^i)}{\pi_{\theta_{\mathrm{old}}}(s_n^i\mid q,s_{<n}^i)}.
\end{equation}

This segment-level credit assignment ensures that the gradient signal is
applied only to the effective segments that are retained and eventually
contribute to the final reasoning trajectory, thereby improving the stability
of reinforcement learning.

\subsection{Erasable Reinforcement Learning}

Given a candidate segment $s_n^i$ from the $i$-th sequence, the non-Markovian
erasure operator $\mathcal{E}$ maps the generation state to a binary decision:
\begin{equation}
\mathcal{E}:
\left(
s_n^i,
U_n^i,
\Theta_{n,e_n^i}
\right)
\mapsto
\{\top,\bot\},
\end{equation}
where $\top$ denotes erasure and $\bot$ denotes acceptance. Based on the comprehensive uncertainty metric $U_n^i$ and the dynamic threshold
$\Theta_{n,e_n^i}$, the erasure trigger is defined as
\begin{equation}
\mathcal{E}(s_n^i)
=
\begin{cases}
\bot,
&
U_n^i \leq \Theta_{n,e_n^i},
\quad \text{accept},
\\
\top,
&
U_n^i > \Theta_{n,e_n^i},
\quad \text{erase}.
\end{cases}
\end{equation}

Let the generation state before producing the $n$-th segment be
$S_n^i=(s_{<n}^i,e_n^i)$, where $e_n^i$ records the number of erasure attempts
for the current segment. The state transition is defined as
\begin{equation}
S_{\mathrm{next}}^i
=
\begin{cases}
\left(
s_{<n}^i \oplus s_n^i,
0
\right),
&
\mathcal{E}(s_n^i)=\bot,
\\
\left(
s_{<n}^i,
e_n^i+1
\right),
&
\mathcal{E}(s_n^i)=\top
\ \land\
e_n^i<E_{\max},
\\
\left(
s_{<n}^i \oplus s_n^i,
0
\right),
&
\mathcal{E}(s_n^i)=\top
\ \land\
e_n^i=E_{\max}.
\end{cases}
\end{equation}

In the erasure branch, the system removes the current pathological segment,
resamples it from the same prefix, and updates the erasure threshold:
\begin{equation}
s_n^i
\sim
\pi_{\theta}
\left(
\cdot
\mid
q,s_{<n}^i
\right),
\qquad
e_n^i
\leftarrow
e_n^i+1,
\qquad
\Theta_{n,e_n^i}
=
\beta^{\mathrm{macro}}_n
\cdot
\gamma_{e_n^i}
\cdot
\phi\left(\mathcal{H}_{<n}\right).
\end{equation}

Therefore, $\mathrm{E}^3\mathrm{RL}$ introduces an explicit rollback-and-retry
mechanism into autoregressive reinforcement learning. Instead of forcing every
sampled segment to be permanently committed, the model dynamically detects,
erases, and regenerates high-uncertainty segments before they propagate into
later reasoning steps.

\section{Experiments}
\label{sec:experiments}


\subsection{Experimental Setup}

\textbf{Datasets and Benchmarks.} We selected 51k samples from the DeepMath-103k\citep{he2025deepmath111} dataset for reinforcement learning training. To rigorously assess the reasoning capabilities of the models, we evaluate the trained policies across a spectrum of highly competitive and complex mathematical benchmarks: AMC 2023\citep{amc2023_12a_aops, amc2023_12b_aops}, AIME 2024\citep{aime2024_i_aops, aime2024_ii_aops}, AIME 2025\citep{aime2025_i_aops, aime2025_ii_aops}, AIME 2026\citep{aime2026_i_aops, aime2026_ii_aops}, MATH 500\citep{lightman2023let}, Minerva\citep{lewkowycz2022solving}, and OlympiadBench\citep{he2024olympiadbench}.

\textbf{Models and Baselines.} We conduct experiments at two parameter scales based on the Qwen3 architecture: Qwen3-4B and Qwen3-8B\citep{yang2025qwen3}.
To contextualize the performance of $\mathrm{E}^3\mathrm{RL}$, we compare it against several strong baselines: Vanilla, GRPO\citep{shao2024deepseekmath111}, DAPO\citep{yu2026dapo111}, GSPO\citep{zheng2025group111}, and SAPO\citep{gao2025soft111}.

\textbf{Evaluation Metrics.} Following standard protocols for mathematical reasoning, we report the \textit{Avg@k} and \textit{Pass@k} metrics, setting $k=32$.

\begin{figure*}[t]
\centering
\includegraphics[width=0.96\linewidth]
{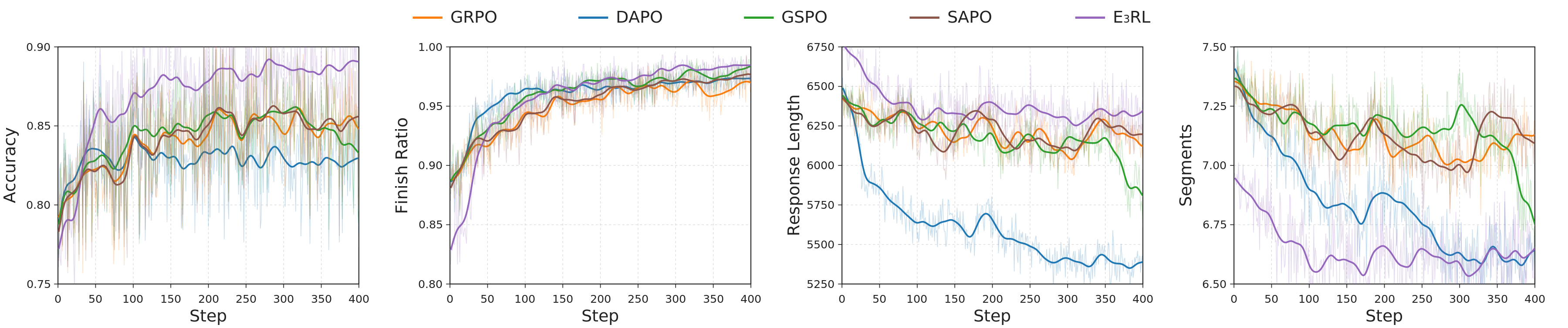}
\caption{Training dynamics of different RL strategies. }
\label{fig:training_dynamics111}
\end{figure*}

\begin{figure*}[t]
\centering
\includegraphics[width=0.96\linewidth]
{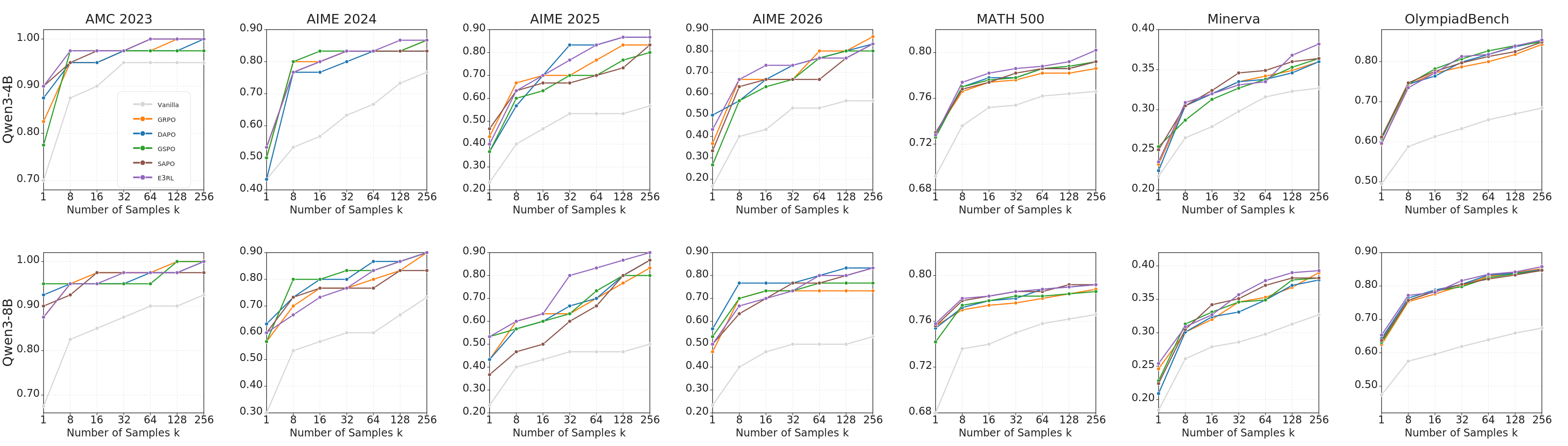}
\caption{Pass@k scaling of different RL strategies. }
\label{fig:pass@k_dynamics111}
\end{figure*}

\begin{table*}[!t]
\centering
\scriptsize
\renewcommand{\arraystretch}{1.2} 
\setlength{\tabcolsep}{3pt} 
\begin{tabular*}{\linewidth}{@{\extracolsep{\fill}}lcccccccccccccc@{}}
\toprule
\multirow{2}{*}{\textbf{Method}}  & \multicolumn{2}{c}{\textbf{AMC 2023}} & \multicolumn{2}{c}{\textbf{AIME 2024}} & \multicolumn{2}{c}{\textbf{AIME 2025}} & \multicolumn{2}{c}{\textbf{AIME 2026}} & \multicolumn{2}{c}{\textbf{MATH 500}} & \multicolumn{2}{c}{\textbf{Minerva}} & \multicolumn{2}{c}{\textbf{OlympiadBench}} \\
\cmidrule(lr){2-3}\cmidrule(lr){4-5}\cmidrule(lr){6-7}\cmidrule(lr){8-9}\cmidrule(lr){10-11}\cmidrule(lr){12-13}\cmidrule(lr){14-15}
& \textit{Avg} &\textit{Pass} & \textit{Avg} &\textit{Pass}&\textit{Avg}&\textit{Pass}&\textit{Avg}&\textit{Pass}&\textit{Avg}&\textit{Pass}&\textit{Avg}&\textit{Pass}&\textit{Avg}&\textit{Pass} \\
\hline
\multicolumn{15}{l}{\textbf{Qwen3-4b}}\\
\addlinespace[0.2em]
Vanilla      & 0.712 & 0.900 & 0.358 & 0.633 & 0.229 & 0.533 & 0.254 & 0.533 & 0.685 & 0.754 & 0.204 & 0.298 & 0.490 & 0.633 \\
GRPO  & 0.829 & \textbf{0.975} & 0.442 & \textbf{0.833} & 0.375 & 0.700 & 0.433 & 0.667 & 0.727 & 0.776 & 0.235 & 0.335 & 0.601 & 0.787 \\
DAPO     & 0.842 & \textbf{0.975} & 0.436 & 0.800 & 0.359 & \textbf{0.833} & 0.419 & \textbf{0.733} & 0.725 & 0.778 & 0.238 & 0.335 & 0.606 & 0.799 \\
GSPO & 0.849 & \textbf{0.975} & 0.445 & \textbf{0.833} & 0.382 & 0.700 & 0.417 & 0.667 & 0.731 & 0.778 & 0.241 & 0.327 & 0.618 & 0.807 \\
SAPO  & 0.855 & \textbf{0.975} & 0.475 & \textbf{0.833} & 0.374 & 0.667 & 0.445 & 0.667 & 0.733 & 0.782 & 0.242 & \textbf{0.346} & 0.615 & 0.797 \\
$\text{E}^3\text{RL}$   & \textbf{0.878} & \textbf{0.975} & \textbf{0.506} & \textbf{0.833} & \textbf{0.433} & 0.767 & \textbf{0.485} & \textbf{0.733} & \textbf{0.742} & \textbf{0.786} & \textbf{0.256} & 0.331 & \textbf{0.639} & \textbf{0.813} \\
\hline
\multicolumn{15}{l}{\textbf{Qwen3-8b}}\\
\addlinespace[0.2em]
Vanilla      & 0.670 & 0.875 & 0.349 & 0.600 & 0.215 & 0.467 & 0.253 & 0.533 & 0.668 & 0.750 & 0.195 & 0.286 & 0.472 & 0.619 \\
GRPO  & 0.831 & \textbf{0.975} & 0.519 & 0.767 & 0.399 & 0.633 & 0.458 & 0.733 & 0.737 & 0.776 & 0.240 & 0.346 & 0.629 & 0.801 \\
DAPO         & 0.846 & 0.950 & 0.538 & 0.800 & 0.414 & 0.667 & 0.508 & \textbf{0.767} & 0.738 & 0.780 & 0.239 & 0.331 & 0.637 & 0.805 \\
GSPO   & 0.842 & 0.950 & 0.521 & \textbf{0.833} & 0.409 & 0.633 & 0.482 & 0.733 & 0.745 & 0.782 & 0.241 & 0.346 & 0.636 & 0.798 \\
SAPO     & 0.843 & \textbf{0.975} & 0.507 & 0.767 & 0.394 & 0.600 & 0.466 & \textbf{0.767} & 0.743 & \textbf{0.786} & 0.246 & 0.351 & 0.639 & 0.805 \\
$\text{E}^3\text{RL}$ & \textbf{0.887} & \textbf{0.975} & \textbf{0.575} & 0.767 & \textbf{0.448} & \textbf{0.800} & \textbf{0.513} & 0.733 & \textbf{0.758} & \textbf{0.786} & \textbf{0.253} & \textbf{0.357} & \textbf{0.654} & \textbf{0.816} \\
\bottomrule
\end{tabular*}
\caption{Avg@32 and Pass@32 results of different RL strategies.}
\label{tab:main_result1}
\end{table*}

\subsection{Main Results}

Table~\ref{tab:main_result1} summarizes the main results of different reinforcement learning strategies on seven mathematical reasoning benchmarks. Overall, $\text{E}^3\text{RL}$ achieves the strongest \textit{Avg@32} performance at both model scales, demonstrating that dynamic epistemic entropy guided erasure consistently improves the average quality of long-horizon reasoning trajectories.

\textbf{Training Dynamics.} As shown in Figure~\ref{fig:training_dynamics111}, $\text{E}^3\text{RL}$ achieves the highest training accuracy while maintaining a competitive finish ratio. It produces longer responses with fewer reasoning segments, suggesting that the model learns to preserve useful prefixes and erase only unstable local fragments. This behavior confirms that entropy-orchestrated erasure improves exploration quality and stabilizes long-range credit assignment.

\textbf{Pass@k Scaling.} Figure~\ref{fig:pass@k_dynamics111} shows that $\text{E}^3\text{RL}$ delivers consistently strong Pass@k curves across both Qwen3-4B and Qwen3-8B. The advantage is especially visible on AIME, Minerva, and OlympiadBench, where increasing samples leads to more reliable success. These results indicate that $\text{E}^3\text{RL}$ improves not only average trajectory quality but also multi-sample recoverability.

\subsection{Ablation Experiment}

\begin{table}[H]
\centering
\captionsetup{skip=2pt}
\scriptsize
\renewcommand{\arraystretch}{1.08} 
\setlength{\tabcolsep}{1pt} 
\begin{tabular*}{\linewidth}{@{\extracolsep{\fill}}lccccccc@{}}
\toprule
\textbf{Method} & \textbf{AMC 2023} & \textbf{AIME 2024} & \textbf{AIME 2025} & \textbf{AIME 2026} & \textbf{MATH 500} & \textbf{Minerva} & \textbf{OlympiadBench} \\
\midrule
\multicolumn{8}{l}{\textbf{Qwen3-4b}}\\
\addlinespace[0.2em]
$\text{E}^3\text{RL}$           & \textbf{0.878} & 0.506 & \textbf{0.433} & \textbf{0.485} & \textbf{0.742} & \textbf{0.256} & \textbf{0.639} \\
w/o extremum deviation          & 0.870 & \textbf{0.512} & 0.425 & 0.477 & 0.738 & 0.252 & 0.634 \\
w/o gradient anomaly            & 0.867 & 0.498 & 0.427 & 0.474 & 0.735 & 0.251 & 0.631 \\
w/o base uncertainty            & 0.841 & 0.445 & 0.430 & 0.426 & 0.723 & 0.236 & 0.605 \\
ow base uncertainty             & 0.868 & 0.502 & 0.431 & 0.481 & 0.737 & 0.254 & 0.636 \\
ow gradient anomaly             & 0.839 & 0.443 & 0.432 & 0.432 & 0.726 & 0.242 & 0.607 \\
ow extremum deviation           & 0.834 & 0.450 & 0.413 & 0.421 & 0.724 & 0.238 & 0.599 \\
\bottomrule
\end{tabular*}
\caption{Ablation study on cognitive entropy components evaluated on the Avg@32 metric. "w/o" represent "with out" while "ow" for "only with".}
\label{tab:avg32_result_entropy}
\end{table}
\vspace{-0.7em}

\begin{table}[H]
\centering
\captionsetup{skip=2pt}
\scriptsize
\renewcommand{\arraystretch}{1.08} 
\setlength{\tabcolsep}{1pt} 
\begin{tabular*}{\linewidth}{@{\extracolsep{\fill}}lccccccc@{}}
\toprule
\textbf{Method} & \textbf{AMC 2023} & \textbf{AIME 2024} & \textbf{AIME 2025} & \textbf{AIME 2026} & \textbf{MATH 500} & \textbf{Minerva} & \textbf{OlympiadBench} \\
\midrule 
\multicolumn{8}{l}{\textbf{Qwen3-4B}}\\
\addlinespace[0.2em]
$\text{E}^3\text{RL}$       & \textbf{0.878} & \textbf{0.506} & \textbf{0.433} & \textbf{0.485} & \textbf{0.742} & \textbf{0.256} & \textbf{0.639} \\
w/o frequency penalty       & 0.872 & 0.503 & 0.429 & 0.481 & 0.739 & 0.255 & 0.637 \\
w/o causal allocation       & 0.865 & 0.494 & 0.416 & 0.476 & 0.737 & 0.252 & 0.634 \\
w/o group dynamics          & 0.854 & 0.486 & 0.398 & 0.468 & 0.733 & 0.249 & 0.625 \\
ow group dynamics           & 0.863 & 0.492 & 0.412 & 0.473 & 0.740 & 0.253 & 0.632 \\
ow causal allocation        & 0.855 & 0.483 & 0.394 & 0.459 & 0.734 & 0.248 & 0.628 \\
ow frequency penalty        & 0.846 & 0.467 & 0.386 & 0.431 & 0.729 & 0.239 & 0.617 \\
\bottomrule
\end{tabular*}
\caption{Ablation study on system mechanisms evaluated on the Avg@32 metric. "w/o" represent "with out" while "ow" for "only with".}
\label{tab:avg32_results_system}
\end{table}
\vspace{-0.8em}

To deeply understand the inner workings of $\text{E}^3\text{RL}$ and validate our theoretical design, we conduct extensive ablation studies on the Qwen3-4B model. We decompose the system into two primary dimensions: the components of the cognitive uncertainty metric and the structural system mechanisms. We evaluate these variations using the rigorous \textit{Avg@32} metric to observe their impact on the expected stability of the reasoning trajectories.

\textbf{Deconstructing Epistemic Entropy Monitoring.} 
Table~\ref{tab:avg32_result_entropy} isolates the effects of the three distinct uncertainty signals: base uncertainty ($\widetilde{\mathcal{H}}_n$), gradient anomaly ($\Delta\mathcal{H}_n$), and extremum deviation ($\mathcal{H}_n^{\max}$). The results clearly indicate that the \textit{base uncertainty} acts as the foundational pillar of the erasure mechanism. Removing it (\textit{w/o base uncertainty}) triggers catastrophic performance drops across all benchmarks, most notably plummeting from $0.485$ to $0.426$ on AIME 2026, and from $0.506$ to $0.445$ on AIME 2024. Conversely, relying exclusively on base uncertainty (\textit{ow base uncertainty}) retains a significant portion of the model's performance but fails to reach the optimal state. The gradient anomaly and extremum deviation components act as critical high-frequency filters. Relying on them alone (\textit{ow gradient anomaly} or \textit{ow extremum deviation}) yields the poorest results, proving they are insufficient as standalone indicators of logical collapse. However, when integrated into the full $\text{E}^3\text{RL}$ system, they effectively detect local burst-like cognitive crises and sharp intra-segment oscillations, pushing the performance to its peak. 

\textbf{Dissecting Structural System Mechanisms.}
Table~\ref{tab:avg32_results_system} isolates the impact of our non-Markovian system designs: the micro-level frequency penalty ($\Gamma(e_n)$), segment-level causal allocation, and the macro-level group dynamics ($\beta_n^{\mathrm{macro}}$). The most severe degradation occurs when the system operates only with frequency penalty (\textit{ow frequency penalty}), dropping AIME 2024 to $0.467$ and AIME 2025 to $0.386$. This strongly supports our theoretical claim that without adaptive group dynamics to scale thresholds based on problem ambiguity, static thresholding fails to generalize across varying difficulty levels. Furthermore, the removal of the frequency penalty (\textit{w/o frequency penalty}) slightly degrades overall performance. This confirms its designed role: the exponentially increasing micro-penalty prevents the model from falling into endless resampling loops on highly difficult segments, successfully precluding reasoning deadlocks. Finally, the ablation of causal allocation (\textit{w/o causal allocation}) results in a broad performance decay, validating that reconstructing the gradient flow pathways to exclusively reward successfully retained logic segments fundamentally preserves the multi-scale credit assignment system.

\begin{figure}[H]
\centering
\includegraphics[width=0.96\linewidth]
{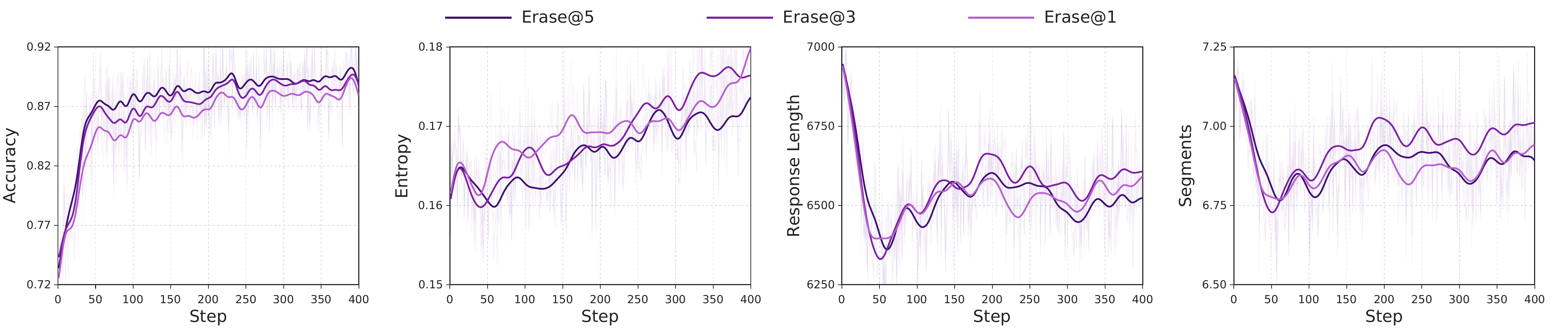}
\caption{Training dynamics of $\text{E}^3\text{RL}$ under different Erase@k settings.}
\label{fig:training_dynamics222}
\end{figure}

\begin{table}[H]
\centering
\scriptsize
\renewcommand{\arraystretch}{1.2} 
\setlength{\tabcolsep}{1pt} 
\begin{tabular*}{\linewidth}{@{\extracolsep{\fill}}lccccccc@{}}
\toprule
\textbf{Method} & \textbf{AMC 2023} & \textbf{AIME 2024} & \textbf{AIME 2025} & \textbf{AIME 2026} & \textbf{MATH 500} & \textbf{Minerva} & \textbf{OlympiadBench} \\
\midrule 
\multicolumn{8}{l}{\textbf{Qwen3-8B}}\\
\addlinespace[0.2em]
erase@1           & 0.855 & 0.543 & 0.419 & 0.484 & 0.747 & 0.239 & 0.642 \\
erase@3        & 0.874 & 0.559 & 0.426 & 0.501 & 0.752 & 0.248 & 0.647 \\
erase@5        & \textbf{0.887} & \textbf{0.575} & \textbf{0.448} & \textbf{0.513} & \textbf{0.758} & \textbf{0.253} & \textbf{0.654} \\
\bottomrule
\end{tabular*}
\caption{Erase@k results of $\text{E}^3\text{RL}$, evaluated using Avg@32.}
\label{tab:avg32_results_system222}
\end{table}

\begin{table}[H]
\centering
\scriptsize
\renewcommand{\arraystretch}{1.2} 
\setlength{\tabcolsep}{1pt} 
\begin{tabular*}{\linewidth}{@{\extracolsep{\fill}}lccccccc@{}}
\toprule
\textbf{Method} & \textbf{AMC 2023} & \textbf{AIME 2024} & \textbf{AIME 2025} & \textbf{AIME 2026} & \textbf{MATH 500} & \textbf{Minerva} & \textbf{OlympiadBench} \\
\midrule 
\multicolumn{8}{l}{\textbf{Qwen3-8B}}\\
\addlinespace[0.2em]
32 $\times$ 256           & 0.851 & 0.544 & 0.433 & 0.489 & 0.748 & 0.246 & 0.643 \\
16 $\times$ 512        & \textbf{0.896} & \textbf{0.583} & \textbf{0.452} & \textbf{0.525} & \textbf{0.761} & \textbf{0.254} & \textbf{0.662} \\
8 $\times$ 1024        & 0.887 & 0.575 & 0.448 & 0.513 & 0.758 & 0.253 & 0.654 \\
4 $\times$ 2048        & 0.875 & 0.568 & 0.441 & 0.506 & 0.750 & 0.249 & 0.649 \\
2 $\times$ 4096        & 0.842 & 0.543 & 0.423 & 0.484 & 0.744 & 0.245 & 0.638 \\

\bottomrule
\end{tabular*}
\caption{Study on the segments $\times$ length configuration of $\text{E}^3\text{RL}$, evaluated using Avg@32.}
\label{tab:avg32_results_system333}
\end{table}

\section{Analysis}
\label{sec:analysis}

\begin{figure*}[t]
\centering
\includegraphics[width=0.98\linewidth]
{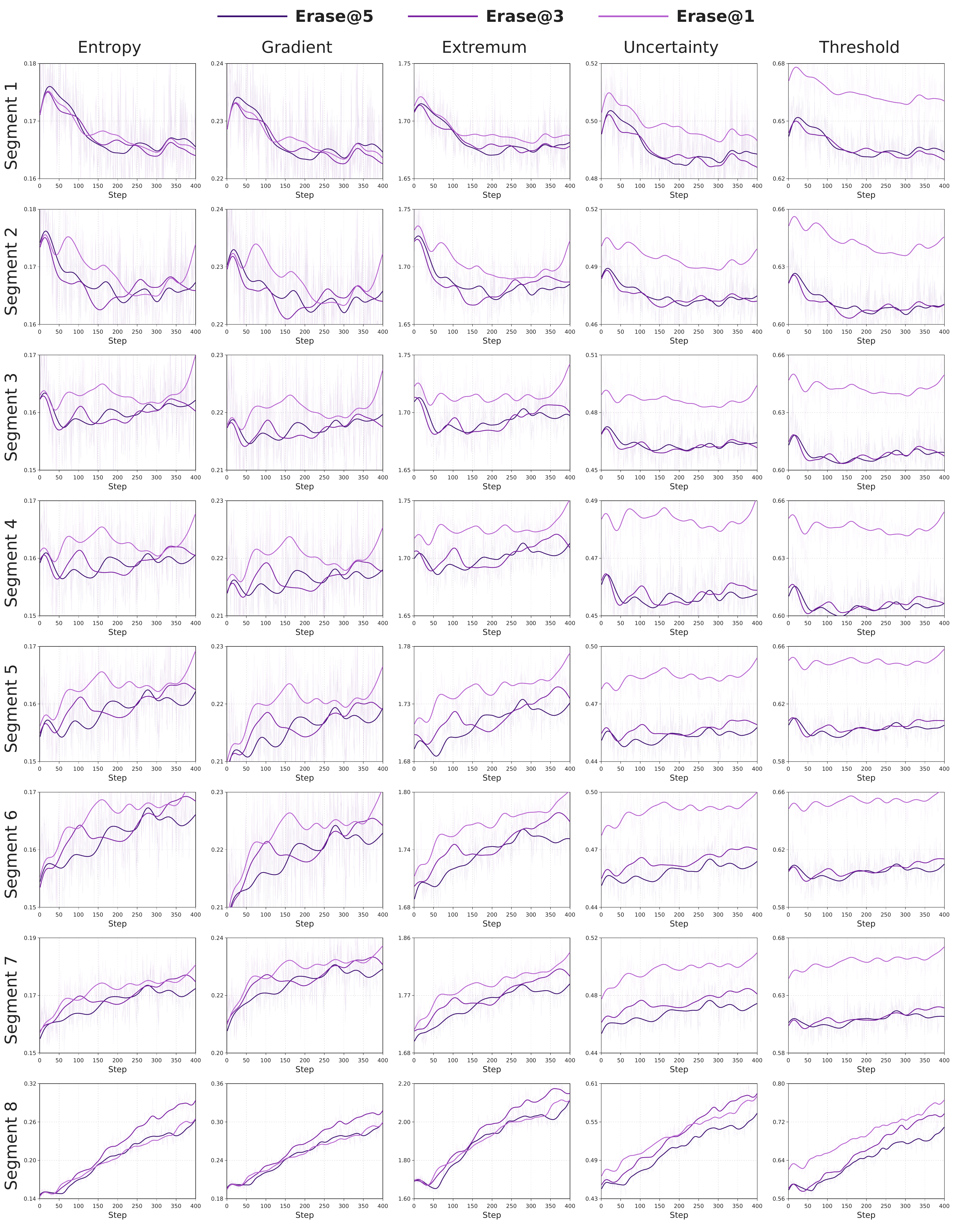}
\caption{Segment-level training dynamics of $\text{E}^3\text{RL}$ under different Erase@k settings.}
\label{fig:training_dynamics666}
\end{figure*}

To further investigate the behavior of E$^3$RL, we analyze two hyperparameters that directly control the granularity of self-correction: the maximum number of erasure attempts and the segment-length configuration. The analysis focuses on Qwen3-8B and follows the same Avg@32 evaluation protocol used in the main experiments.  Together, Figure~\ref{fig:training_dynamics222}, Table~\ref{tab:avg32_results_system222}, Table~\ref{tab:avg32_results_system333}, and Figure~\ref{fig:training_dynamics666} show that E$^3$RL benefits from a sufficiently large erasure budget and a balanced segmentation scheme, while overly restrictive rollback or overly coarse segmentation weakens local correction.

\textbf{Erase@k Scaling.} As shown in Figure~\ref{fig:training_dynamics222}, larger Erase@k budgets produce more stable training dynamics. Erase@5 achieves the highest accuracy curve and maintains smoother improvement than Erase@1, indicating that allowing multiple localized retries gives the model a broader search space before committing an uncertain segment. Table~\ref{tab:avg32_results_system222} provides quantitative confirmation: Erase@5 obtains the best Avg@32 on all seven benchmarks, improving over Erase@1 from 0.855 to 0.887 on AMC 2023, from 0.543 to 0.575 on AIME 2024, and from 0.642 to 0.654 on OlympiadBench. The monotonic trend from Erase@1 to Erase@3 and Erase@5 suggests that repeated erasure mainly helps difficult reasoning problems where early local defects are more likely to induce downstream collapse.


\textbf{Segment-Granularity Trade-off.} Table~\ref{tab:avg32_results_system333} shows that segmentation itself has a non-trivial optimum. The 16 $\times$ 512 configuration achieves the best Avg@32 across all benchmarks, reaching 0.896 on AMC 2023, 0.583 on AIME 2024, 0.525 on AIME 2026, and 0.662 on OlympiadBench. In contrast, 32 $\times$ 256 underperforms because overly short segments may fragment coherent reasoning and trigger excessive local decisions, while 2 $\times$ 4096 performs worst because long segments delay error detection and resemble ordinary irreversible generation. The default 8 $\times$ 1024 setting remains competitive, but the 16 $\times$ 512 result indicates that finer yet still semantically meaningful checkpoints can better balance correction precision and reasoning continuity.

\textbf{Overall Analysis.} These results indicate that E$^3$RL is not merely benefiting from additional sampling, but from where and how additional computation is spent. Increasing Erase@k improves the ability to repair unstable segments, while the segment-level dynamics in Figure~\ref{fig:training_dynamics666} show that uncertainty is concentrated unevenly across the reasoning trajectory.

\section{Related Work}
\label{sec:related_work}

Recent reinforcement learning methods have substantially advanced the reasoning
capabilities of large language models. GRPO\citep{shao2024deepseekmath}
removes the critic model used in PPO and estimates advantages from group-level
relative rewards, reducing the training cost of RL for reasoning models.
DAPO\citep{yu2025dapo} further improves large-scale RL training through
decoupled clipping, dynamic sampling, token-level policy-gradient loss, and
overlong reward shaping, alleviating entropy collapse and improving training
stability. GSPO\citep{zheng2025group} replaces token-level importance ratios
with sequence-level likelihood ratios, making policy optimization more stable
and better aligned with sequence-level rewards. SAPO\citep{gao2025soft}
introduces soft adaptive policy optimization by replacing hard clipping with
smooth temperature-controlled scaling, enabling more stable and informative
policy updates. BAPO\citep{xi2025bapo} studies off-policy RL for LLMs and uses
balanced adaptive clipping to preserve policy entropy and stabilize optimization
under stale-policy data. DisCO\citep{li2025disco} reformulates reasoning RL as
discriminative constrained optimization, mitigating the question-level
difficulty bias of group-relative objectives and improving training stability.

\section{limitation \& future discussion}

Although $\text{E}^3\text{RL}$ achieves consistent gains on long-horizon
mathematical reasoning benchmarks, several aspects remain worth further
exploration. First, our experiments focus on mathematical and logical reasoning,
where long-chain dependency and error propagation are especially prominent. This
setting provides a natural testbed for studying the autoregressive curse, while
future work may further examine the applicability of erasable reinforcement
learning to other structured generation tasks, such as code reasoning,
tool-augmented problem solving, and multi-turn agentic workflows. Second, the current implementation uses predefined segment length, smoothing
window size, and erasure-attempt budget. These choices are lightweight and work
well in our experiments, but more adaptive controllers may further improve
efficiency across different model scales and task distributions. For example,
future systems could dynamically adjust the segment granularity according to the
local uncertainty profile of each reasoning trajectory.

\section{Conclusion}
\label{sec:conclusion}


In this paper, we propose Dynamic Epistemic Entropy Orchestrated Erasable
Reinforcement Learning ($\text{E}^3\text{RL}$), a segment-level reinforcement
learning framework for mitigating error propagation in long-horizon
autoregressive reasoning. Instead of treating generation as a fully irreversible
one-shot trajectory, $\text{E}^3\text{RL}$ introduces an endogenous
uncertainty-driven erasure mechanism that detects high-risk reasoning segments,
rolls them back, and regenerates them before they contaminate subsequent
decisions. By combining epistemic entropy monitoring, adaptive thresholding,
frequency-aware erasure control, and segment-level advantage assignment, the
proposed method provides a lightweight mechanism for local correction without
relying on external process reward models or full-sequence rejection. Experiments on multiple mathematical reasoning benchmarks demonstrate that
$\text{E}^3\text{RL}$ consistently improves reasoning performance across both
Qwen3-4B and Qwen3-8B models. The ablation studies further verify the importance
of the proposed uncertainty metric and system-level erasure mechanisms,
indicating that effective long-horizon reasoning benefits from both reliable
local uncertainty estimation and structurally compatible policy optimization.
Overall, $\text{E}^3\text{RL}$ offers a practical step toward more robust
autoregressive reasoning systems with endogenous error-correction capabilities.

\small
\bibliographystyle{unsrtnat}
\bibliography{references}

\appendix

\newpage

\section{Ethics Statement}
\label{sec:ethics}

This work studies reinforcement learning methods for improving the reasoning
capabilities of large language models, with experiments conducted on
mathematical and logical reasoning benchmarks. The proposed method does not
involve human subjects, private user data, or personally identifiable
information. All training and evaluation data used in this study are drawn from
publicly available or standard research datasets. The primary goal of $\text{E}^3\text{RL}$ is to improve the reliability and
sample efficiency of long-horizon reasoning by reducing the propagation of local
reasoning errors. While stronger reasoning models may provide broad benefits in
education, scientific problem solving, and automated assistance, they may also be
adapted for unintended uses when deployed without appropriate safeguards. As
with other methods that enhance LLM reasoning, practical deployment should be
accompanied by standard safety measures, including content filtering,
misuse monitoring, and human oversight in high-stakes applications. We also note that $\text{E}^3\text{RL}$ operates as a training and inference
mechanism and does not introduce external reward models or additional sources of
user data. Therefore, its ethical considerations are primarily aligned with
those of general-purpose large language models. We encourage future applications
of this framework to follow responsible AI practices, including transparent
evaluation, careful domain-specific validation, and appropriate limitations on
use in sensitive decision-making contexts.

\section{Reproducibility}
\label{sec:reproducibility}
We take several steps to support the reproducibility of our experiments. First,
we clearly specify the model backbones, training data, evaluation benchmarks,
and metrics used throughout the paper. All experiments are conducted on the
Qwen3-4B and Qwen3-8B architectures, trained with 51k samples selected from the
DeepMath-103k dataset, and evaluated on AMC 2023, AIME 2024, AIME 2025,
AIME 2026, MATH 500, Minerva, and OlympiadBench using \textit{Avg@32} and
\textit{Pass@32}. Second, the main algorithmic components of $\text{E}^3\text{RL}$ are described
in Section~\ref{sec:method}, including segmented generation, epistemic entropy
monitoring, adaptive thresholding, erasure control, and segment-level advantage
assignment. The mathematical definitions of the uncertainty metric, dynamic
threshold, erasure operator, and optimization objective are provided to make the
training procedure implementable without relying on unspecified external
modules. Third, all baseline methods are evaluated under the same model scales, training
data, and benchmark protocols to ensure a fair comparison. The ablation studies
in Section~\ref{sec:analysis} further isolate the contribution of each major
component, making it possible to verify the functional role of the proposed
design choices. For full reproducibility, implementation details such as training
hyperparameters, decoding configuration, segment length, smoothing window size,
maximum erasure attempts, batch size, learning rate schedule, and hardware setup
will be provided in the appendix. This will allow future work to reproduce the reported results and
extend $\text{E}^3\text{RL}$ to other reasoning tasks and model scales.

\section{LLM usage}
We partially used large language models (LLMs) exclusively for non-scientific writing assistance, specifically for language polishing, clarity improvement, and suggestions. No parts of the core methodology, experiments, or results were generated by LLMs.

\section{Experimental Setup}
\label{sec:experimental_setup}

\textbf{Datasets and Benchmarks.}
We train all models on 51k samples selected from DeepMath-103k. To evaluate
long-horizon mathematical reasoning ability, we report results on seven
benchmarks: AMC 2023, AIME 2024, AIME 2025, AIME 2026, MATH 500, Minerva, and
OlympiadBench. Following common practice in mathematical reasoning evaluation,
we use \textit{Avg@32} and \textit{Pass@32} as the main metrics.

\textbf{Models and Baselines.}
We conduct experiments on two model scales based on the Qwen3 architecture:
Qwen3-4B and Qwen3-8B. We compare $\text{E}^3\text{RL}$ with several strong RL
baselines, including Vanilla, GRPO, DAPO, GSPO, and SAPO. All methods are trained
and evaluated under the same data split, model backbone, and evaluation
protocols to ensure a fair comparison.

\textbf{Training Configuration.}
All models are trained for 2 epochs with a batch size of 128. We use a learning
rate of $1\times10^{-6}$, a warmup ratio of $0.05$, and a KL regularization
coefficient of $0.04$. The maximum context length is set to 13,312 tokens. For
segmented generation, the segment length is set to 1,024 tokens and the maximum
number of segments is set to 8. During sampling, we use temperature $1.0$,
top-$p=0.9$, and top-$k=50$. Each prompt is sampled with 8 generations during
training.

\textbf{Erasable Generation Configuration.}
For $\text{E}^3\text{RL}$, each reasoning trajectory is generated in a segmented
manner. High-uncertainty segments are detected by the proposed epistemic entropy
metric and can be erased and regenerated before being committed to the prefix.
The maximum number of erasure attempts for each segment is set to 5. The
frequency penalty is applied to discourage excessive repeated erasures, while
the macro-level group dynamics adaptively calibrates the erasure threshold
according to the uncertainty distribution of sampled trajectories.

\textbf{Infrastructure and Evaluation.}
Training is conducted on 32 NVIDIA H100 GPUs. We adopt a distributed
training--rollout setup, where rollout servers are used for efficient generation
and synchronized with the training worker during RL optimization. DeepSpeed
ZeRO-2 is used for distributed training. We evaluate the model every 50 training
steps and save checkpoints at the same interval. The final reported results are
computed using the same decoding and evaluation protocol across all compared
methods.

\section{Complexity Analysis}
\label{sec:time_complexity}

We analyze the computational cost of $\text{E}^3\text{RL}$ and compare it with
standard GRPO. Let $T$ denote the maximum generation length, $G$ the number of
sampled responses per prompt, $d$ the hidden dimension, $M$ the number of
Transformer layers, $|\mathcal{V}|$ the vocabulary size, $\ell$ the segment
length, and $N=\lceil T/\ell\rceil$ the number of segments.

\textbf{Complexity of GRPO.}
In standard GRPO, each prompt is associated with $G$ sampled trajectories. The
dominant computation comes from autoregressive generation and policy-gradient
optimization over the generated sequences. For a Transformer language model, the
per-sequence computation can be written as
\begin{equation}
C_{\mathrm{LM}}(T)
=
\mathcal{O}\left(
M T^2 d + M T d^2
\right),
\end{equation}
where the first term corresponds to self-attention and the second term
corresponds to feed-forward and projection operations. Therefore, the per-update
complexity of GRPO is
\begin{equation}
C_{\mathrm{GRPO}}
=
\mathcal{O}\left(
G M T^2 d + G M T d^2
\right).
\end{equation}
When the hidden dimension is large, the feed-forward term often dominates, but
we keep both terms for completeness.

\textbf{Additional cost of $\text{E}^3\text{RL}$.}
Compared with GRPO, $\text{E}^3\text{RL}$ introduces three additional
computational components. First, epistemic entropy is computed from the output
distribution. Since logits are already produced during generation, this only
requires an additional reduction over the vocabulary:
\begin{equation}
C_{\mathrm{ent}}
=
\mathcal{O}\left(GT|\mathcal{V}|\right).
\end{equation}
Second, segment-level statistics, smoothing, and threshold comparison are
computed once per segment:
\begin{equation}
C_{\mathrm{seg}}
=
\mathcal{O}\left(GNW\right),
\end{equation}
where $W$ is the smoothing window size. Since $N=T/\ell$, this term is linear in
the number of generated tokens.

Third, $\text{E}^3\text{RL}$ may erase and regenerate high-uncertainty
segments. Let $e$ denote the expected total number of segment regenerations per
sampled trajectory. The corresponding expected regenerated length is
$\ell e$, and the relative regeneration ratio is
\begin{equation}
r
=
\frac{\ell e}{T}.
\end{equation}
The extra language-model computation caused by erasure is therefore
\begin{equation}
C_{\mathrm{erase}}
=
\mathcal{O}\left(
rG M T^2 d + rG M T d^2
\right).
\end{equation}

Combining the above terms, the total expected complexity of $\text{E}^3\text{RL}$
is
\begin{equation}
\begin{aligned}
C_{\text{E}^3\text{RL}}
=
\mathcal{O}\Big(
&
(1+r)G M T^2 d
+
(1+r)G M T d^2
\\
&
+
GT|\mathcal{V}|
+
GNW
\Big).
\end{aligned}
\end{equation}
Equivalently, relative to GRPO, we have
\begin{equation}
\frac{
C_{\text{E}^3\text{RL}}
}{
C_{\mathrm{GRPO}}
}
=
1+r
+
\mathcal{O}\left(
\frac{
|\mathcal{V}| + W/\ell
}{
M(Td+d^2)
}
\right).
\end{equation}

This shows that $\text{E}^3\text{RL}$ preserves the same asymptotic order as
GRPO when the expected regeneration ratio $r$ is bounded. The additional entropy
monitoring and thresholding operations are lightweight compared with the
Transformer forward and backward computations, while the main extra cost comes
from the regenerated segments. Since erasure is performed locally at the segment
level rather than by discarding complete trajectories, the overhead scales with
the amount of regenerated content instead of the full sequence length.

\textbf{Memory footprint.}
$\text{E}^3\text{RL}$ maintains only one active segmented trajectory for each
sampled response. When a segment is erased, the method rolls back to the cached
prefix state and regenerates the current segment without maintaining a branching
search tree. Therefore, its rollout memory usage remains linear in the sequence
length:
\begin{equation}
\mathcal{M}_{\text{E}^3\text{RL}}
=
\mathcal{O}\left(GMTd\right)
+
\mathcal{O}\left(GN\right),
\end{equation}
where the first term corresponds to KV-cache storage and the second term
corresponds to segment-level statistics. This contrasts with tree-based
reasoning methods, whose memory may grow with the number of maintained branches.
Thus, $\text{E}^3\text{RL}$ introduces local correction while retaining a linear
memory footprint.

\begin{table*}[!t]
\centering
\small
\renewcommand{\arraystretch}{1.2} 
\setlength{\tabcolsep}{3.5pt}
\resizebox{\textwidth}{!}{%
\begin{tabular}{@{}l*{7}{C}*{7}{C}@{}}
\toprule
\multirow{3}{*}{\textbf{Method}} & \multicolumn{14}{c}{\textbf{Pass@k}} \\
\cmidrule(lr){2-15}
& \multicolumn{7}{c}{\textbf{Qwen3-4B}} & \multicolumn{7}{c}{\textbf{Qwen3-8B}} \\
\cmidrule(lr){2-8} \cmidrule(lr){9-15}
& $k{=}1$ & $k{=}8$ & $k{=}16$ & $k{=}32$ & $k{=}64$ & $k{=}128$ & $k{=}256$ & $k{=}1$ & $k{=}8$ & $k{=}16$ & $k{=}32$ & $k{=}64$ & $k{=}128$ & $k{=}256$ \\
\midrule

\multicolumn{15}{l}{\textbf{AMC 2023}} \\
Vanilla      & 0.700 & 0.875 & 0.900 & 0.950 & 0.950 & 0.950 & 0.950 & 0.675 & 0.825 & 0.850 & 0.875 & 0.900 & 0.900 & 0.925 \\
GRPO         & 0.825 & 0.950 & 0.950 & \textbf{0.975} & 0.975 & \textbf{1.000} & \textbf{1.000} & 0.875 & \textbf{0.950} & \textbf{0.975} & \textbf{0.975} & \textbf{0.975} & \textbf{1.000} & \textbf{1.000} \\
DAPO         & 0.875 & 0.950 & 0.950 & \textbf{0.975} & 0.975 & 0.975 & \textbf{1.000} & 0.925 & \textbf{0.950} & 0.950 & 0.950 & \textbf{0.975} & 0.975 & \textbf{1.000} \\
GSPO         & 0.775 & \textbf{0.975} & \textbf{0.975} & \textbf{0.975} & 0.975 & 0.975 & 0.975 & \textbf{0.950} & \textbf{0.950} & 0.950 & 0.950 & 0.950 & \textbf{1.000} & \textbf{1.000} \\
SAPO         & \textbf{0.900} & 0.950 & \textbf{0.975} & \textbf{0.975} & \textbf{1.000} & \textbf{1.000} & \textbf{1.000} & 0.900 & 0.925 & \textbf{0.975} & \textbf{0.975} & \textbf{0.975} & 0.975 & 0.975 \\
$\text{E}^3\text{RL}$ & \textbf{0.900} & \textbf{0.975} & \textbf{0.975} & \textbf{0.975} & \textbf{1.000} & \textbf{1.000} & \textbf{1.000} & 0.875 & \textbf{0.950} & 0.950 & \textbf{0.975} & \textbf{0.975} & 0.975 & \textbf{1.000} \\
\midrule

\multicolumn{15}{l}{\textbf{AIME 2024}} \\
Vanilla      & 0.433 & 0.533 & 0.567 & 0.633 & 0.667 & 0.733 & 0.767 & 0.300 & 0.533 & 0.567 & 0.600 & 0.600 & 0.667 & 0.733 \\
GRPO         & 0.500 & \textbf{0.800} & 0.800 & \textbf{0.833} & \textbf{0.833} & 0.833 & \textbf{0.867} & 0.567 & 0.700 & 0.767 & 0.767 & 0.800 & 0.833 & \textbf{0.900} \\
DAPO         & 0.433 & 0.767 & 0.767 & 0.800 & \textbf{0.833} & 0.833 & 0.833 & \textbf{0.633} & 0.733 & \textbf{0.800} & 0.800 & \textbf{0.867} & \textbf{0.867} & \textbf{0.900} \\
GSPO         & 0.500 & \textbf{0.800} & \textbf{0.833} & \textbf{0.833} & \textbf{0.833} & 0.833 & \textbf{0.867} & 0.567 & \textbf{0.800} & \textbf{0.800} & \textbf{0.833} & 0.833 & \textbf{0.867} & \textbf{0.900} \\
SAPO         & \textbf{0.533} & 0.767 & 0.800 & \textbf{0.833} & \textbf{0.833} & 0.833 & 0.833 & 0.600 & 0.733 & 0.767 & 0.767 & 0.767 & 0.833 & 0.833 \\
$\text{E}^3\text{RL}$ & \textbf{0.533} & 0.767 & 0.800 & \textbf{0.833} & \textbf{0.833} & \textbf{0.867} & \textbf{0.867} & 0.600 & 0.667 & 0.733 & 0.767 & 0.833 & \textbf{0.867} & \textbf{0.900} \\
\midrule

\multicolumn{15}{l}{\textbf{AIME 2025}} \\
Vanilla      & 0.233 & 0.400 & 0.467 & 0.533 & 0.533 & 0.533 & 0.567 & 0.233 & 0.400 & 0.433 & 0.467 & 0.467 & 0.467 & 0.500 \\
GRPO         & 0.433 & \textbf{0.667} & \textbf{0.700} & 0.700 & 0.767 & 0.833 & 0.833 & 0.433 & \textbf{0.600} & \textbf{0.633} & 0.633 & 0.700 & 0.767 & 0.833 \\
DAPO         & 0.367 & 0.567 & \textbf{0.700} & \textbf{0.833} & \textbf{0.833} & \textbf{0.867} & \textbf{0.867} & 0.433 & 0.567 & 0.600 & 0.667 & 0.700 & 0.800 & 0.867 \\
GSPO         & 0.367 & 0.600 & 0.633 & 0.700 & 0.700 & 0.767 & 0.800 & \textbf{0.533} & 0.567 & 0.600 & 0.633 & 0.733 & 0.800 & 0.800 \\
SAPO         & \textbf{0.467} & 0.633 & 0.667 & 0.667 & 0.700 & 0.733 & 0.833 & 0.367 & 0.467 & 0.500 & 0.600 & 0.667 & 0.800 & 0.867 \\
$\text{E}^3\text{RL}$ & 0.400 & 0.633 & \textbf{0.700} & 0.767 & \textbf{0.833} & \textbf{0.867} & \textbf{0.867} & \textbf{0.533} & \textbf{0.600} & \textbf{0.633} & \textbf{0.800} & \textbf{0.833} & \textbf{0.867} & \textbf{0.900} \\
\midrule

\multicolumn{15}{l}{\textbf{AIME 2026}} \\
Vanilla      & 0.167 & 0.400 & 0.433 & 0.533 & 0.533 & 0.567 & 0.567 & 0.233 & 0.400 & 0.467 & 0.500 & 0.500 & 0.500 & 0.533 \\
GRPO         & 0.367 & \textbf{0.667} & 0.667 & 0.667 & \textbf{0.800} & \textbf{0.800} & \textbf{0.867} & 0.467 & 0.700 & 0.733 & 0.733 & 0.733 & 0.733 & 0.733 \\
DAPO         & \textbf{0.500} & 0.567 & 0.667 & \textbf{0.733} & 0.767 & \textbf{0.800} & 0.833 & \textbf{0.567} & \textbf{0.767} & \textbf{0.767} & \textbf{0.767} & \textbf{0.800} & \textbf{0.833} & \textbf{0.833} \\
GSPO         & 0.267 & 0.567 & 0.633 & 0.667 & 0.767 & \textbf{0.800} & 0.800 & 0.533 & 0.700 & 0.733 & 0.733 & 0.767 & 0.767 & 0.767 \\
SAPO         & 0.333 & 0.633 & 0.667 & 0.667 & 0.667 & 0.767 & 0.833 & 0.500 & 0.633 & 0.700 & \textbf{0.767} & 0.767 & 0.800 & \textbf{0.833} \\
$\text{E}^3\text{RL}$ & 0.433 & \textbf{0.667} & \textbf{0.733} & \textbf{0.733} & 0.767 & 0.767 & 0.833 & 0.500 & 0.667 & 0.700 & 0.733 & \textbf{0.800} & 0.800 & \textbf{0.833} \\
\midrule

\multicolumn{15}{l}{\textbf{MATH 500}} \\
Vanilla      & 0.692 & 0.736 & 0.752 & 0.754 & 0.762 & 0.764 & 0.766 & 0.680 & 0.736 & 0.740 & 0.750 & 0.758 & 0.762 & 0.766 \\
GRPO         & \textbf{0.730} & 0.766 & 0.774 & 0.776 & 0.782 & 0.782 & 0.786 & 0.756 & 0.770 & 0.774 & 0.776 & 0.780 & 0.784 & 0.788 \\
DAPO         & 0.728 & 0.770 & 0.778 & 0.778 & 0.786 & 0.786 & 0.792 & 0.754 & 0.772 & 0.778 & 0.780 & \textbf{0.788} & 0.790 & \textbf{0.792} \\
GSPO         & 0.726 & 0.770 & 0.776 & 0.778 & 0.786 & 0.788 & 0.792 & 0.742 & 0.774 & 0.778 & 0.782 & 0.782 & 0.784 & 0.786 \\
SAPO         & \textbf{0.730} & 0.768 & 0.774 & 0.782 & 0.786 & 0.786 & 0.792 & 0.756 & 0.778 & \textbf{0.782} & \textbf{0.786} & 0.786 & \textbf{0.792} & \textbf{0.792} \\
$\text{E}^3\text{RL}$ & 0.728 & \textbf{0.774} & \textbf{0.782} & \textbf{0.786} & \textbf{0.788} & \textbf{0.792} & \textbf{0.802} & \textbf{0.758} & \textbf{0.780} & \textbf{0.782} & \textbf{0.786} & \textbf{0.788} & 0.790 & \textbf{0.792} \\
\midrule

\multicolumn{15}{l}{\textbf{Minerva}} \\
Vanilla      & 0.217 & 0.265 & 0.279 & 0.298 & 0.316 & 0.323 & 0.327 & 0.184 & 0.261 & 0.279 & 0.286 & 0.298 & 0.313 & 0.327 \\
GRPO         & 0.232 & 0.305 & 0.320 & 0.335 & 0.342 & 0.349 & 0.360 & 0.246 & 0.301 & 0.320 & 0.346 & 0.353 & 0.368 & 0.390 \\
DAPO         & 0.224 & 0.305 & 0.320 & 0.335 & 0.338 & 0.346 & 0.360 & 0.209 & 0.301 & 0.324 & 0.331 & 0.349 & 0.371 & 0.379 \\
GSPO         & \textbf{0.254} & 0.287 & 0.313 & 0.327 & 0.338 & 0.353 & 0.364 & 0.228 & \textbf{0.313} & 0.331 & 0.346 & 0.349 & 0.379 & 0.382 \\
SAPO         & 0.250 & 0.305 & \textbf{0.324} & \textbf{0.346} & \textbf{0.349} & 0.360 & 0.364 & 0.224 & 0.305 & \textbf{0.342} & 0.351 & 0.371 & 0.382 & 0.382 \\
$\text{E}^3\text{RL}$ & 0.235 & \textbf{0.309} & 0.320 & 0.331 & 0.335 & \textbf{0.368} & \textbf{0.382} & \textbf{0.254} & 0.309 & 0.327 & \textbf{0.357} & \textbf{0.378} & \textbf{0.390} & \textbf{0.393} \\
\midrule

\multicolumn{15}{l}{\textbf{OlympiadBench}} \\
Vanilla      & 0.495 & 0.588 & 0.613 & 0.633 & 0.655 & 0.670 & 0.684 & 0.474 & 0.575 & 0.596 & 0.619 & 0.639 & 0.659 & 0.674 \\
GRPO         & 0.594 & 0.742 & 0.772 & 0.787 & 0.800 & 0.818 & 0.843 & 0.625 & 0.753 & 0.776 & 0.801 & 0.829 & 0.840 & 0.852 \\
DAPO         & 0.607 & 0.744 & 0.764 & 0.799 & 0.818 & 0.837 & 0.851 & 0.643 & 0.764 & \textbf{0.788} & 0.805 & 0.833 & 0.838 & 0.849 \\
GSPO         & 0.609 & 0.744 & \textbf{0.782} & 0.807 & \textbf{0.827} & \textbf{0.840} & 0.849 & 0.631 & 0.757 & 0.785 & 0.798 & 0.825 & 0.837 & 0.846 \\
SAPO         & \textbf{0.612} & \textbf{0.747} & 0.776 & 0.797 & 0.813 & 0.825 & 0.849 & 0.637 & 0.757 & 0.784 & 0.805 & 0.821 & 0.833 & 0.847 \\
$\text{E}^3\text{RL}$ & 0.596 & 0.735 & 0.772 & \textbf{0.813} & 0.818 & 0.839 & \textbf{0.854} & \textbf{0.653} & \textbf{0.772} & 0.781 & \textbf{0.816} & \textbf{0.835} & \textbf{0.842} & \textbf{0.858} \\
\bottomrule
\end{tabular}%
}
\caption{Comprehensive Pass@k results for Qwen3-4B and Qwen3-8B across different benchmarks and values of $k$.}
\label{tab:pass_at_k_merged}
\end{table*}

\clearpage
\begin{table*}[p]
\centering
\scriptsize
\renewcommand{\arraystretch}{1.2} 
\setlength{\tabcolsep}{1pt} 
\begin{tabular*}{\linewidth}{@{\extracolsep{\fill}}lccccccc@{}}
\toprule
\textbf{Method} & \textbf{AMC 2023} & \textbf{AIME 2024} & \textbf{AIME 2025} & \textbf{AIME 2026} & \textbf{MATH 500} & \textbf{Minerva} & \textbf{OlympiadBench} \\
\midrule
\multicolumn{8}{l}{\textbf{Qwen3-4B}}\\
\addlinespace[0.2em]
$\text{E}^3\text{RL}$           & \textbf{0.975} & \textbf{0.833} & 0.767 & \textbf{0.733} & \textbf{0.786} & 0.331 & \textbf{0.813} \\
w/o extremum deviation          & \textbf{0.975} & -     & \textbf{0.800} & 0.700 & 0.782 & 0.327 & \textbf{0.813} \\
w/o gradient anomaly            & 0.950 & 0.800 & 0.733 & 0.667 & 0.784 & 0.331 & 0.807 \\
w/o base uncertainty            & 0.950 & 0.767 & 0.767 & 0.700 & 0.776 & \textbf{0.335} & 0.804 \\
ow base uncertainty             & \textbf{0.975} & \textbf{0.833} & 0.767 & \textbf{0.733} & 0.782 & 0.329 & 0.811 \\
ow gradient anomaly             & 0.950 & 0.767 & 0.733 & 0.700 & 0.778 & 0.327 & 0.807 \\
ow extremum deviation           & 0.950 & 0.767 & 0.733 & \textbf{0.733} & 0.780 & 0.329 & 0.793 \\
\bottomrule
\end{tabular*}
\caption{Ablation study on cognitive entropy components evaluated on the Pass@32 metric for Qwen3-4B.}
\label{tab:pass32_result_entropy}
\end{table*}
\clearpage

\begin{table*}[p]
\centering
\scriptsize
\renewcommand{\arraystretch}{1.2} 
\setlength{\tabcolsep}{1pt} 
\begin{tabular*}{\linewidth}{@{\extracolsep{\fill}}lccccccc@{}}
\toprule
\textbf{Method} & \textbf{AMC 2023} & \textbf{AIME 2024} & \textbf{AIME 2025} & \textbf{AIME 2026} & \textbf{MATH 500} & \textbf{Minerva} & \textbf{OlympiadBench} \\
\midrule
\multicolumn{8}{l}{\textbf{Qwen3-4B}}\\
\addlinespace[0.2em]
$\text{E}^3\text{RL}$           & \textbf{0.975} & \textbf{0.833} & 0.767 & \textbf{0.733} & \textbf{0.786} & 0.331 & \textbf{0.813} \\
w/o frequency penalty           & 0.950 & \textbf{0.833} & \textbf{0.800} & \textbf{0.733} & 0.784 & \textbf{0.335} & 0.809 \\
w/o causal allocation           & \textbf{0.975} & \textbf{0.833} & 0.767 & 0.700 & 0.784 & 0.327 & 0.809 \\
w/o group dynamics              & \textbf{0.975} & 0.800 & 0.767 & 0.667 & 0.778 & 0.331 & 0.805 \\
ow group dynamics               & \textbf{0.975} & 0.800 & 0.767 & 0.667 & 0.782 & 0.327 & 0.805 \\
ow causal allocation            & \textbf{0.975} & \textbf{0.833} & \textbf{0.800} & 0.700 & 0.782 & 0.324 & 0.801 \\
ow frequency penalty            & 0.950 & 0.800 & 0.767 & 0.667 & 0.776 & 0.327 & 0.803 \\
\bottomrule
\end{tabular*}
\caption{Ablation study on system mechanisms evaluated on the Pass@32 metric for Qwen3-4B.}
\label{tab:pass32_results_system}
\end{table*}
\clearpage





\clearpage

\end{document}